\documentclass[times, twoside]{zHenriquesLab-StyleBioRxiv_mod}
\usepackage{placeins}

\leadauthor{Wilt}

\begin{document}

\title{Scatterbrained: A flexible and expandable pattern for decentralized machine learning} 
\shorttitle{Scatterbrained: Decentralized Federated Learning}

\author[a]{Miller Wilt}
\author[a,b]{Jordan K. Matelsky}
\author[a]{Andrew S. Gearhart}

\affil[a]{Johns Hopkins Applied Physics Laboratory, Laurel, Maryland, United States}
\affil[b]{University of Pennsylvania Department of Bioengineering, Philadelphia, Pennsylvania, United States}

\maketitle

\begin{abstract}
Federated machine learning is a technique for training a model across multiple devices without exchanging data between them. Because data remains local to each compute node, federated learning is well-suited for use-cases in fields where data is carefully controlled, such as medicine, or in domains with bandwidth constraints. One weakness of this approach is that most federated learning tools rely upon a central server to perform workload delegation and to produce a single shared model. Here, we suggest a flexible framework for decentralizing the federated learning pattern, and provide an open-source, reference implementation compatible with PyTorch. 
\end{abstract}

\begin{keywords}
federated learning | distributed computing | edge computing | decentralized
\end{keywords}

\begin{corrauthor}
jordan.matelsky@jhuapl.edu
\end{corrauthor}

\section*{Introduction}

Conventional machine learning technologies have achieved wide adoption due to their high accuracy and adaptability to diverse use-cases. To improve efficiency and minimize cost, data and machine learning tools are most commonly collocated, and neural networks are trained on large, centralized compute resources. This approach is popular, but is poorly suited to certain types of tasks, such as when a dataset contains sensitive data that cannot be shared due to logistical or legal restrictions (e.g., medical data), or in cases where it is impractical to transmit data to centralized resources (e.g., in large-scale edge computing applications, or in bandwidth-austere settings).

To address this problem, researchers have proposed \textit{federated learning} (FL), a paradigm in which individual compute nodes train on their data locally, rather than sending data to a centralized server~\cite{mcmahan2017communication}. Neural network parameters (or weights) --- rather than raw data --- are then shared with a central server, which returns a combined model to each of the edge nodes. In this way, all nodes benefit from their peers' training, without direct access to their peers' data. Such a method is most famously used by the Google GBoard predictive text feature on mobile phone keyboards~\cite{mcmahan2017gboard}.

One major limitation of the traditional federated learning approach is that it requires continuous and secure access to a trusted central server~\cite{ziller2021pysyft}. Many implementations, such as TensorFlow Federated~\cite{tensorflow2015-whitepaper} or PySyft~\cite{ziller2021pysyft}, make similar assumptions about node reliability, and leave it to the developer to resolve server trust as well as node dropout or drop-in connectivity behavior. Consistent communication with a remote compute node can not be guaranteed in all federated learning circumstances, and is almost certainly not the case in a decentralized FL environment, where the number of connections between nodes scales rapidly with the size of the cluster~\cite{hu2019decentralized,khan2021federated,lalitha2019peer,roy2019braintorrent}.

\begin{figure}[t]
    \centering
    \includegraphics[width=0.45\linewidth]{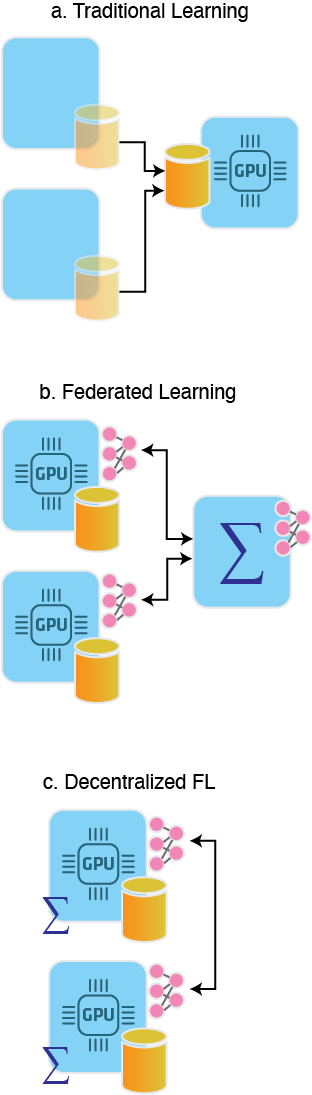}
    \caption{\textbf{a. Traditional learning.} Edge nodes upload raw data to a central server. The central server performs all computation. \textbf{b. Federated learning.} Edge nodes perform local computation and training. A central server performs parameter pooling. \textbf{c. Decentralized federated learning.} Edge nodes perform local computation and training. Each node performs pooling with its networked neighbors.}
    \label{fig:three}
\end{figure}

It is also critical that the cooperative learning algorithm be entirely decoupled from the machine learning model in question. In other words, the representation of the decentralized compute graph should have no bearing upon the model itself, and any machine learning model should be compatible with a decentralized learning framework. We enforce this requirement because we assert that adding a decentralized federated learning regime to an existing tool should not meaningfully impact the model or presentation of data. We illustrate the main differences between traditional machine learning, federated learning, and decentralized federated learning in \textbf{Fig.~\ref{fig:three}}.

\begin{figure}[t!]
    \centering
    \includegraphics[width=\linewidth]{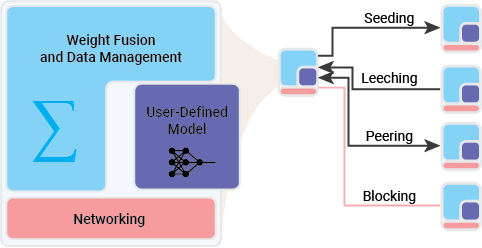}
    \caption{\textbf{A proposed architecture for a decentralized machine learning node.} At left, a single node is composed of: (1) a networking layer, to communicate with other compute nodes in the community, (2) a weight combination and data management component to choose parameter-sharing behavior and train a local model, and (3) a ``pluggable'' module in which a developer can add their machine learning model. At right, the node participates in four different types of network behavior: A ``seeding'' relationship in which the node sends but does not receive weights; a ``leeching'' relationship in which the node receives but does not send weights; a ``peering'' relationship in which the node both sends as well as receives; and a ``blocking'' relationship in which the node performs no communication with other nodes.}
    \label{fig:arch}
\end{figure}

In this work, we suggest critical features for a successful decentralized federated learning platform and share our reference implementation, \textit{Scatterbrained}. Our framework satisfies the requirements of (1) allowing a developer to define use-case specific compute graph topologies, (2) enabling population- or node-level parameter-sharing behavior, (3) permitting node dropout or lossiness, and (4) decoupling machine learning from network communication (\textbf{Fig.~\ref{fig:arch}}).

\section*{Desired Traits of a Decentralized Learning Framework}

\subsection*{Existing implementations of decentralized federated learning}

There are several competing proposals of decentralized federated learning algorithms.~\cite{lalitha2019peer,roy2019braintorrent,daily2018gossipgrad,jiang2017collaborative} Most notably, we call attention to \textit{Lalitha et al.}~\cite{lalitha2019peer} and the \textit{BrainTorrent} algorithm~\cite{roy2019braintorrent}, two distinct approaches to cooperative learning.

Importantly, these existing implementations are purpose-built, and do not afford the end-user sufficient flexibility to manipulate compute-graph topology or node-level data-sharing preferences. It is our belief that these properties should be developer-facing parameters, as different use-cases may call for different graph topology or data sharing strategies. That is, a universal federated learning framework should support all federated learning algorithms discussed in this manuscript, \textit{including} centralized approaches.

\subsection*{Critical features for a federated learning framework}

We propose that there is a set of necessary, adjustable parameters that must be present for a federated learning framework to be useful to the general machine learning community:

\subsubsection*{Compute-graph topology}
\begin{figure}[t]
    \centering
    \includegraphics[width=\linewidth]{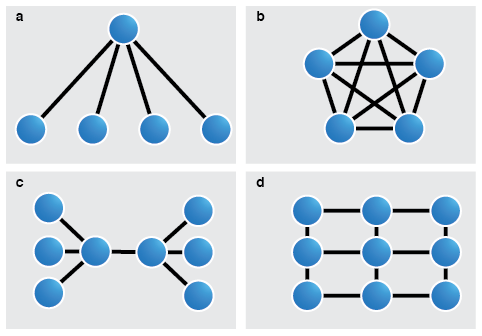}
    \caption{\textbf{Examples of federated learning compute-graph topology.} We propose that a federated learning framework must support diverse compute-graph topologies. \textbf{a. Traditional federated learning.} An example of a centralized federated learning approach, with a central node responsible for data aggregation. \textbf{b. Fully-connected graph.} A fully connected graph enables each node to communicate with every other node, increasing bandwidth usage but maximizing the speed at which new information traverses the graph. \textbf{c. Local neighborhood graph.} Such a graph might be useful when many edge devices connect to a smaller number of higher-performance servers, where bandwidth and compute is inexpensive. \textbf{d. Grid layout.} A grid layout may be useful for geographically distributed nodes in order to improve bandwidth to peers while reducing construction cost.}
    \label{fig:topology}
\end{figure}

The compute graph topology describes how nodes in a federated learning cluster connect to one another. In one extreme, nodes can connect to only a single (central) node: This describes a traditional, centralized federated learning algorithm (\textbf{Fig.~\ref{fig:topology}a}). In the other extreme, each node may connect to every other node in the community, forming a fully connected graph (\textbf{Fig.~\ref{fig:topology}b}). Depending on the use case, it may also be useful for clusters to adopt other connectivity patterns (\textbf{Fig.~\ref{fig:topology}c,d}).

\subsubsection*{Parameter sharing preferences} 

Node-level parameter-sharing preferences refer to how individual nodes participate in cooperative learning. Each node may act as a \textit{read-only}, \textit{write-only}, or \textit{read-and-write} resource in the graph. It is also useful to allow a node's behavior to vary over time based upon internal or external stimuli. For example, a node that has not encountered local data may start in \textit{read-only} mode, accepting inputs from peers until it reaches a threshold volume of useful local data. After this node has its own useful new information to contribute to the network, it may switch into \textit{read-and-write} mode. Other use-cases may necessitate other state transitions between these strategies.

\subsubsection*{Ephemeral nodes}

In distributed edge applications, node network connectivity may be limited, leading to inconsistent node connectivity or permanent dropout. This effect is compounded when the size of the cluster approaches hundreds or thousands of compute nodes. To support applications where individual node network connectivity may be limited, a federated cluster must be robust to node dropout and intermittent bandwidth constraints. This is one of the primary components missing from most existing implementations of decentralized federated learning, and it is one that is most important to guarantee the resiliency of the network as a whole.

This is particularly important for Internet-of-Things devices (IoT), where network access is not only not guaranteed, but is sometimes deliberately withheld to improve energy efficiency~\cite{khan2021federated}.

\subsubsection*{Abstraction of the model from federation} 

We assert that the separation of the machine learning model from the federated learning business logic is a requirement for a successful decentralized federated learning framework. This is because tight integration of a model with the framework imparts an engineering burden on the developer--likely unacceptable for most machine learning applications. Further, by abstracting the machine learning model from federation logic, we enable use of a diverse set of learning tools, such as different ML libraries or different strategies for combining node-level models.

\subsection*{Reference Implementation}

We have developed a Python library, \textit{Scatterbrained}, which abstracts many of the implementation details of networking while enabling a developer to leverage the flexibility of the decentralized learning paradigm. In this work, we demonstrate Scatterbrained compatibility with the popular PyTorch~\cite{pytorch} library, but we believe it is a straightforward extension to extend this work for compatibility with other machine learning libraries.

To maximize the usability of the Scatterbrained library for existing use-cases, we developed a simple API which abstracts low-level node communication across the compute graph and enables fine-grained customization of a node's parameter-sharing behavior. To perform low-overhead communication between Scatterbrained nodes and provide a developer with the ability to design a custom compute graph, we built a communication layer utilizing the 0MQ library~\cite{hintjens2013zeromq}.

\begin{figure*}[th!!!]
    \centering
    \includegraphics[width=\linewidth]{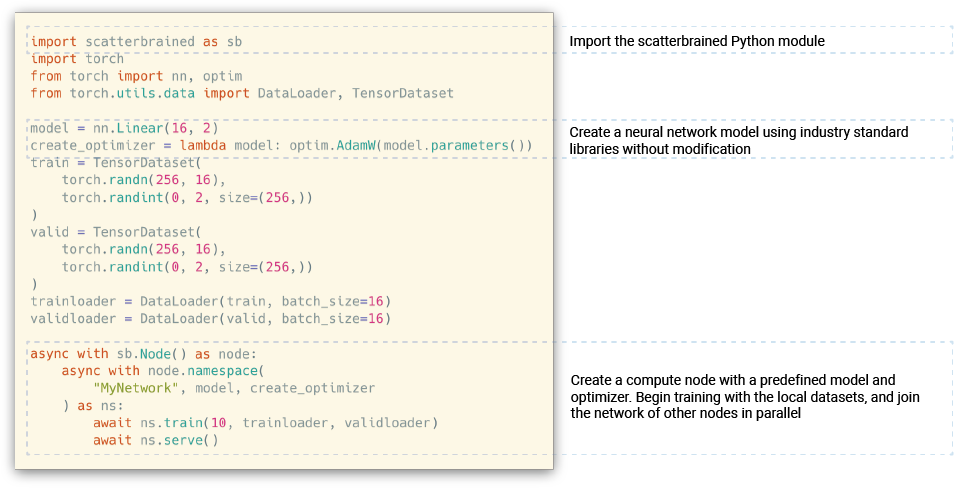} \\
    \includegraphics[width=\linewidth]{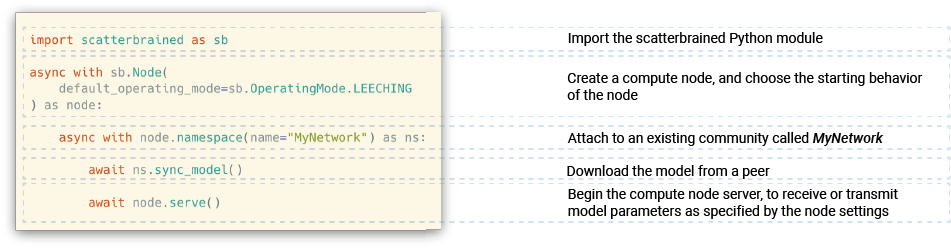}
    \caption{\textbf{A simple example using the \textit{Scatterbrained} library.} In this example, we demonstrate Python code to start a new \textit{Scatterbrained} compute node. The first example demonstrates the creation of a new node with a simple neural network implemented in Torch. The second example illustrates a new node joining the existing federated community by referencing its namespace, ``MyNetwork.'' It then connects in ``leeching'' mode and downloads the model (if it's available) from a peer. This means that the newly connected node does not need to have a local copy of the neural network to begin participating in the federated learning community.}
    \label{fig:example}
\end{figure*}

This high-level networking API performs communication on behalf of the model training and parameter-pooling modules. Our intention is for all \textit{Scatterbrained} operation to be largely independent of the behavior of the ML model itself, and so all communication with the model itself is performed through high-level stateless communication. An architecture diagram of this framework is shown in \textbf{Fig.~\ref{fig:arch}}. To illustrate the simple high-level interface that we propose here, we share example code in \textbf{Fig.~\ref{fig:example}}. In this example, a \textit{Scatterbrained} compute node connects to an existing federated cluster by referencing it by its name. This example shows the power of a federated learning: A compute node can join a community of peers and begin learning with a handful of lines of code.


\section*{Discussion}

Decentralized learning is a desirable property in many domains, including medicine, reconnaissance and exploration, and learning in resource-austere settings. Decentralization enables compute nodes to learn both independently as well as cooperatively, leveraging community knowledge without excessive data sharing. To take advantage of this emerging technology, decentralized learning algorithms will likely vary in both compute-graph connectivity as well as in the sharing behavior of individual nodes. We propose that a general, high-level, federated learning framework can satisfy both of these needs while improving codebase consistency and readability.

In this work, we shared our reference implementation, \textit{Scatterbrained}, a PyTorch-compatible federated learning framework that is applicable to both centralized and decentralized learning use-cases. We expect that this open-source software package will enable the community to more rapidly explore the rich space of decentralized federated learning algorithms.

Because the field of decentralized learning is nascent, there remain many considerations and challenges to consider before cooperative learning tools are commonplace.

\subsection*{Data Privacy}

Despite the inherent privacy associated with sharing model parameters rather than complete datasets, it is an open question as to whether decentralized federated learning satisfies the requirements of complete data security. In other words, it is still unknown whether an adversary in the federated community could successfully reverse-engineer training data from the transmitted weights of peers alone. Though this property is common with traditional machine learning~\cite{hua2018reverse}, data privacy is of particular interest in the federated learning community. 

We propose that selective node connectivity and weight-sharing behavior is sufficient to ameliorate the ability of an adversary to reverse-engineer training data from a federated learning peer (e.g., avoiding sending and receiving weights to/from the same partner in the same epoch), but further work will be required to demonstrate this to a sufficient degree to trust such tools indiscriminately with sensitive data. Similarly, noise might be introduced to the weights prior to transmission: This will reduce the likelihood of data leakage, but at the expense of performance and collective accuracy~\cite{naseri2021local}.

\subsection*{Compute-Graph Optimization}

We leave the option of compute-graph optimizations available to the users of the \textit{Scatterbrained} library. We believe that there is a rich opportunity to optimize the graph of node connectivity alongside the training process to improve measures such as the rate of information-gain across the community of nodes or the sequestering of adversarial nodes. Though this work is outside of the scope of this manuscript, the \textit{Scatterbrained} networking layer supports this behavior, thanks to the flexible architecture described above.

\subsection*{Future Directions}

Decentralized federated learning is a fertile domain for future research, and it is likely that low-power, low-bandwidth edge-computing resources will be key players in the future of distributed computing. As we look to the future of federated learning, we anticipate that distributed, decentralized approaches will enable new and increasingly efficient computing paradigms to address challenges throughout the machine learning world.

\section*{Supplemental Materials}

All code and scripts referenced in this manuscript are available at \hyperlink{https://github.com/JHUAPL/scatterbrained}{github.com/JHUAPL/scatterbrained}.

\begin{acknowledgements}

This work was completed with the support of JHU/APL Internal Research Funding.

\end{acknowledgements}

\FloatBarrier
\section*{Bibliography}
\bibliography{bibliography}

\onecolumn
\newpage

\end{document}